# A Unifying Framework for Structural Properties of CSPs: Definitions, Complexity, Tractability

**Lucas Bordeaux** lucasb@microsoft.com
*Microsoft Research*
*7 J J Thomson Avenue*
*Cambridge, CB3 0FB, United Kingdom*

**Marco Cadoli** cadoli@dis.uniroma1.it
**Toni Mancini** tmancini@dis.uniroma1.it
*Dipartimento di Informatica e Sistemistica*
*Sapienza Università di Roma*
*Via Ariosto 25, I-00185 Roma, Italy*

## Abstract

Literature on Constraint Satisfaction exhibits the definition of several "structural" properties that can be possessed by CSPs, like (in)consistency, substitutability or interchangeability. Current tools for constraint solving typically detect such properties efficiently by means of incomplete yet effective algorithms, and use them to reduce the search space and boost search.

In this paper, we provide a *unifying framework* encompassing most of the properties known so far, both in CSP and other fields' literature, and shed light on the semantical relationships among them. This gives a unified and comprehensive view of the topic, allows new, unknown, properties to emerge, and clarifies the computational complexity of the various detection problems.

In particular, among the others, two new concepts, *fixability* and *removability* emerge, that come out to be the *ideal* characterisations of values that may be safely assigned or removed from a variable's domain, while preserving problem satisfiability. These two notions subsume a large number of known properties, including inconsistency, substitutability and others.

Because of the computational intractability of all the property-detection problems, by following the CSP approach we then determine a number of relaxations which provide sufficient conditions for their tractability. In particular, we exploit forms of *language restrictions* and *local reasoning*.

## 1. Introduction

Many Constraint Satisfaction Problems (CSPs) which arise in the modelling of real-life applications exhibit "structural" properties that distinguish them from random instances. Detecting such properties has been widely recognised to be an effective way for improving the solving process. To this end, several of them have already been identified, and different techniques have been developed in order to exploit them, with the goal of reducing the search space to be explored. Good examples are value inconsistency (Mackworth, 1977; Montanari, 1974), substitutability and interchangeability (Freuder, 1991), more general forms of symmetries (Crawford, Ginsberg, Luks, & Roy, 1996; Gent & Smith, 2000), and functional dependencies among variables (Li, 2000; Mancini & Cadoli, 2007).

Unfortunately, checking whether such properties hold, is (or is thought to be) often computationally hard. As an example, let us consider interchangeability. Value $a$ is said to be interchangeable with value $b$ for variable $x$ if every solution which assigns $a$ to $x$ remains a solution if $x$ is changed to $b$, and vice versa (Freuder, 1991). The problem of checking interchangeability is coNP-complete (cf. Proposition 4). Analogously, detecting some other forms of symmetry reduces to the graph automorphism problem (Crawford, 1992) (for which there is no known polynomial time algorithm, even if there is evidence that it is not NP-complete, Köbler, Schöning, & Torán, 1993).

To this end, in order to allow general algorithms to exploit such properties efficiently, different approaches can be followed. First of all, syntactic restrictions on the constraint languages can be enforced, in order to allow the efficient verification of the properties of interest. Alternatively, "local" versions of such properties can be defined, that can be used to infer their global counterparts, and that can be verified in polynomial time. As an instance of this "local reasoning" approach, instead of checking whether a value is *fully* interchangeable for a variable, Freuder (1991) proposes to check whether that value is *neighbourhood*, or $k$-interchangeable. This task involves considering only bounded-sized subsets of the constraints, and hence can be performed in polynomial time. Neighbourhood and $k$-interchangeability are sufficient (but not necessary) conditions for full interchangeability, and have been proven to be highly effective in practice (cf., e.g., Choueiry & Noubir, 1998; Lal, Choueiry, & Freuder, 2005).

In this paper we give a formal characterisation of several properties of CSPs which can be exploited in order to save search, and present a unifying framework for them that allows for their semantical connections to emerge. Some of the properties are well-known in the Constraint Programming literature, others have been used in other contexts (as in databases), while some others are, to the best of our knowledge, original, and, in our opinion, play a key role in allowing a deep understanding of the topic. In particular, we reconsider the notions of *inconsistency*, *substitutability* and *interchangeability*, and propose those of *fixable*, *removable*, and *implied* value for a given variable, which are instances of the more general definition of *satisfiability-preserving transformation* and those of *determined*, *dependent*, and *irrelevant* variable. These properties make it possible to transform a problem into a simpler one. Depending on the case, this transformation is guaranteed to preserve all solutions, or the satisfiability of the problem, i.e., at least a solution, if one exists. In general, each of these properties can be detected either statically, during a preprocessing stage of the input CSP (cf., e.g., Cadoli & Mancini, 2007), or dynamically, during search (since they may arise at any time). Moreover, in some cases we don't even need to explicitly verify whether some properties hold, because this is guaranteed by the intrinsic characteristics of the problem. For instance, some problems are guaranteed to have a unique solution. Such cases are referred to as "promise problems" in the literature (Even, Selman, & Yacobi, 1984), meaning that in addition to the problem description we are informed of certain properties it verifies (cf. forthcoming Example 1 for an example).

The formal characterisation of the properties and their connections allow us to shed light on the computational complexity of their recognition task in a very elegant way, proving that, in the worst case, the detection of each of them is as complex as the original problem. In particular, as we will see in Section 3.1, detecting each of the proposed properties is a coNP-complete task. This holds also for Freuder's substitutability and interchangeability (this result is, to the best of our knowledge, original). Hence, in order to be able to practically make the relevant checks during preprocessing and search, we investigate two different approaches for the efficient verification of the proposed properties: additions of suitable *restrictions* to the constraint language, and exploitation of efficient forms of *local reasoning*, i.e., by checking them for single constraints.

The outline of the paper is as follows: after giving an intuitive example and recalling some preliminaries, in Section 2 we present the properties we are interested in, and discuss their connections. Then, in Section 3 we focus on the complexity of the various property-detection tasks. In particular, in Section 3.1 we prove that all of them are intractable; hence, in Section 3.2 we focus on relaxations that guarantee tractability of reasoning, investigating the two aforementioned approaches. Finally, in Section 4 we draw conclusions and address future work.

## 2. A Hierarchy of Properties

In this section, we formally define several structural properties of CSPs, discuss the semantical relationships that hold among them, and show how they can be exploited during constraint solving.

### 2.1 An Intuitive Example

In order to allow for a gentle introduction of the main properties investigated in the forthcoming sections, we first introduce the following example.

**Example 1** (Factoring, Lenstra & Lenstra, 1990; Pyhälä, 2004)**.** *This problem is a simplified version of one of the most important problems in public-key cryptography. Given a (large) positive integer $Z$ and the fact that it is a product of two different unknown* prime *numbers $X$ and $Y$ (different from 1), the goal is to identify these two numbers.*

*An intuitive formulation of any instance of this problem (i.e., for any given $Z$) as a CSP, adequate for arbitrarily large numbers, amounts to encode the combinatorial circuit of integer multiplication, and is as follows: assuming $Z$ has $n$ digits (in base $b$) $z_1, \ldots, z_n$, we consider $2n$ variables $x_1, \ldots, x_n$ and $y_1, \ldots, y_n$ one for each digit (in base $b$) of $X$ and $Y$ ($x_1$ and $y_1$ being the least significant ones). The domain for all these variables is $[0, b-1]$. In order to maintain information about the carries, $n+1$ additional variables $c_1, \ldots, c_{n+1}$ must be considered, with domain $[0..(b-1)^2 n/b]$.*[1]

*As for the constraints (cf. Figure 1 for the intuition, where $x_4$, $x_5$, $x_6$, $y_4$, $y_5$, $y_6$ are equal to 0, and are omitted for readability), they are the following:*

1. *Constraints on factors:*

1. In this intuitive example, with a little abuse with respect to what will be permitted by forthcoming Definition 1, we allow, to enhance readability, different variables to be defined over different domains. However, we observe that it is easy to recover from this by using standard and well-known techniques (e.g., adding new monadic constraints to model smaller domains).

| | | | | | | | |
|---|---|---|---|---|---|---|---|
| | | | | 7 | 8 | 7 | * |
| | | | | 7 | 9 | 7 | = |
| 0 | 6 | 13 | 18 | 12 | 4 | 0 | |
| | | | | 49 | 56 | 49 | |
| | | | 63 | 72 | 63 | – | |
| | | 49 | 56 | 49 | – | – | |
| | 6 | 2 | 7 | 2 | 3 | 9 | |
| | | | | $x_3$ | $x_2$ | $x_1$ | * |
| | | | | $y_3$ | $y_2$ | $y_1$ | = |
| $c_7$ | $c_6$ | $c_5$ | $c_4$ | $c_3$ | $c_2$ | $c_1$ | |
| | | | | $x_3y_1$ | $x_2y_1$ | $x_1y_1$ | |
| | | | $x_3y_2$ | $x_2y_2$ | $x_1y_2$ | – | |
| | | $x_3y_3$ | $x_2y_3$ | $x_1y_3$ | – | – | |
| | $z_6$ | $z_5$ | $z_4$ | $z_3$ | $z_2$ | $z_1$ | |

Figure 1: Factoring instance 627239, $n = 6$, $b = 10$

(a) *Factors must be different from 1, or, equivalently,* $X \neq Z$ *and* $Y \neq Z$ *must hold;*

(b) *For every digit* $i \in [1, n]$*:* $z_i = c_i + \sum_{j,k \in [1,n]: j+k=i+1} (x_j * y_k \mod b)$;

2. *Constraints on carries:*

   (a) *Carry on the least significant digit is 0:* $c_1 = 0$;

   (b) *Carries on other digits:* $\forall i \in [2, n+1]$, $c_i = c_{i-1} + \sum_{j,k \in [1,n]: j+k=i} \frac{x_j * y_k}{b}$;

   (c) *Carry on the most significant digit is 0:* $c_{n+1} = 0$.

□

When a guess on the two factors $X$ and $Y$ (i.e., on variables $x_1, \ldots, x_n$ and $y_1, \ldots, y_n$) has been made, values for variables $c_1, \ldots, c_{n+1}$ are completely determined, since they follow from the semantics of the multiplication. This is called a *functional dependence* among variables.

Functional dependencies arise very often in, e.g., problems for which an intermediate state has to be maintained, and their detection and exploitation has been recognized to be of great importance from an efficiency point of view, since it can lead to significant reductions of the search space (cf., e.g., Giunchiglia, Massarotto, & Sebastiani, 1998; Mancini & Cadoli, 2007; Cadoli & Mancini, 2007).

Moreover, the presence of functional dependencies among variables of a CSP highlights a second interesting problem, i.e., that of *computing* the values of dependent variables after a choice of the defining ones has been made. This problem, which is always a subproblem of a CSP with dependencies, has exactly *one* solution, hence, the knowledge of such a *promise* can be useful to the solver. It is worth noting that there are also problems which intrinsically exhibit promises. This is the case of, e.g., Factoring where we add the symmetry-breaking constraint forcing $x_1, \ldots, x_n$ to be lexicographically less than or equal to $y_1, \ldots, y_n$. This new formulation is guaranteed to have exactly one solution (since both $X$ and $Y$ are prime).

The Factoring problem exhibits also other interesting properties: let us consider an instance such that $Z$ is given in binary notation (i.e., $b = 2$) and with the least significant digit $z_1$ being equal to 1. This implies that the last digit of both factors $X$ and $Y$ must be 1. Hence, we can say that value 1 is *implied* for variables $x_1$ and $y_1$, and that 0 is *removable* for them and, more precisely *inconsistent*. Moreover, for this problem, which, if the symmetry is broken, has a unique solution, we also know that all variables $x_1, \ldots, x_n$ and $y_1, \ldots, y_n$ are *determined* (cf. forthcoming Definition 2), regardless of the instance, and because of the functional dependence already discussed, we have that variables encoding carries, i.e., $c_i$ ($i \in [1, n]$), are *dependent* on $\{x_1, \ldots, x_n, y_1, \ldots, y_n\}$.

As for problems with unique solutions, it is known that, unfortunately, their resolution remains intractable (cf., e.g., Papadimitriou, 1994; Valiant & Vijay V. Vazirani, 1986; Calabro, Impagliazzo, Kabanets, & Paturi, 2003). However, this does not exclude the possibility to find good heuristics for instances with such a promise, or to look for other properties that are implied by the existence of unique solutions, that can be exploited in order to improve the search process. In particular, determined and implied values play an important role in this and other classes of problems. As the Factoring example shows, such problems arise frequently in practice, either as subproblems of other CSPs, as in presence of functional dependencies (cf. also Mancini & Cadoli, 2007; Mancini, Cadoli, Micaletto, & Patrizi, 2008, for more examples), or because of intrinsic characteristics of the problem at hand. In general, if a problem has a unique solution, all variables have a determined value.

Another central role is played by the removability property, that characterises precisely the case when a value can be safely removed from the domain of a variable, while preserving satisfiability. This property is of course weaker than inconsistency (since some solutions may be lost), but can be safely used in place of it when we are interested in finding only a solution of a CSP, if one exists, and not all of them.

### 2.2 Preliminaries

**Definition 1** (Constraint Satisfaction Problem (CSP), Dechter, 1992)**.** *A Constraint Satisfaction Problem is a triple $\langle X, D, C\rangle$ where:*

- *$X$ is a finite set of* variables*;*
- *$D$ is a finite set of values, representing the* domain *for each variable;*
- *$C$ is a finite set of* constraints *$\{c_1, \ldots c_{|C|}\}$, with each of them being of the form $c_i = r_i(V_i)$, where $V_i$ is a list of $k \leq |X|$ variables in $X$ (the constraint scope), and $r_i \subseteq D^k$ is a $k$-ary relation over $D$ (the constraint relation). We sometimes will denote the set $V_i$ of variables of constraint $c_i$ by $var(c_i)$.*

□

Given a set of variables $V$ and a domain $D$, a $V$-tuple $t$ over $D$ is a mapping which associates a value $t_x \in D$ to every $x \in V$. Such value is called the *$x$-component* of $t$. Given a $V$-tuple $t$ and a subset $U \subseteq V$ of its variables, we denote by $t|_U$ the *restriction* of $t$ to $U$, which has the same value as $t$ on the variables of $U$ and is undefined elsewhere. The explicit *assignment* of the value $a \in D$ to the $x$-component of a $V$-tuple $t$ ($x \in V$) is written $t[x := a]$.

Given a CSP $\langle X, D, C\rangle$, an $X$-tuple $t$ *satisfies* a constraint $c_i = r_i(V_i) \in C$ if $t|_{V_i} \in r_i$. We denote by $Sol(c_i)$ the set of $X$-tuples which satisfy $c_i$. The set $\bigcap_{c \in C} Sol(c)$ of $X$-tuples which satisfy all the constraints is called the *solution space*, and denoted $Sol(C)$.

By solving a CSP we mean to decide whether the set $Sol(C)$ is non-empty and, if so, compute one (or all) the solutions.

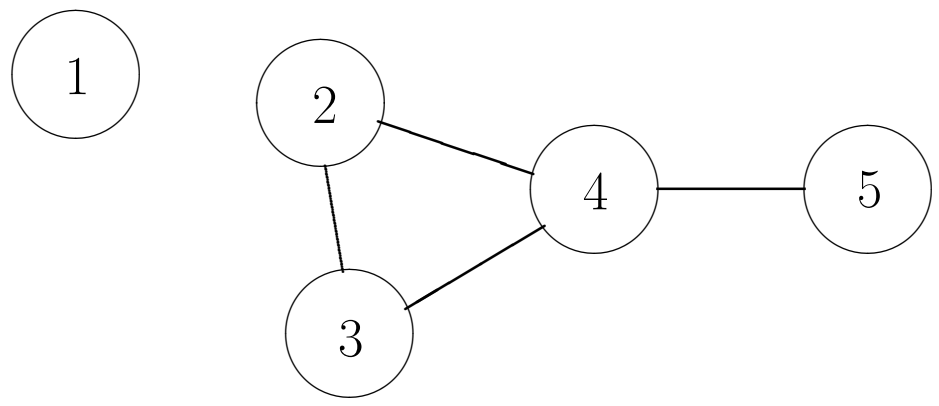

Figure 2: A graph to be 3-colored

The set of $X$-tuples $t$ over $D$ is called the *search space* and denoted by $S_D$, or simply $S$ if the domain is implicit from the context. The relational operators of *selection*, *projection* and *complement* will be useful: given a $V$-relation $c$, a subset $U$ of $V$ and a value $a \in D$, we denote by $\sigma_{x=a}(c_i)$ (resp. $\sigma_{x \neq a}(c_i)$) the $V$-relation which contains the tuples of $c_i$ whose value on $x$ is $a$ (resp. is different from $a$), by $\pi_U(c_i)$ the set of restrictions to $U$ of tuples of $c_i$ (i.e., the set of $U$-tuples $\{t \mid \exists t' \in c_i \ (t = t'|_U)\}$) and by $\overline{c_i}$ the set of $V$-tuples $\{t \mid t \notin c_i\}$.

Note that $\sigma_{x=a}(S)$ denotes the search space obtained by fixing the value of $x$ to $a$. For the sake of simplicity, the sets $X$ and $C$ will be considered as globally defined and shall therefore be omitted from the parameters of most definitions; only the search space will be explicitly mentioned.

**Example 2.** *Consider the CSP $\langle X, D, C \rangle$ modeling the 3-coloring problem for the graph in Figure 2. We have that:*

- $X = \{x_1, \dots, x_5\}$ *is the set of variables (one for each node),*
- $D = \{R, G, B\}$ *is the set of colors, and*
- $C$ *is the following finite set of constraints, one for each edge:*

$$C = \{NE(x_2, x_3), NE(x_3, x_4), NE(x_2, x_4), NE(x_4, x_5)\},$$

*where $NE$ (not-equal) is a binary relation defined as*

$$(\{R, G, B\} \times \{R, G, B\}) \ \setminus \ \{\langle R, R \rangle, \langle G, G \rangle, \langle B, B \rangle\}.$$

### 2.3 Definitions

In this section, we formally present the properties already introduced in Section 2.1, and show their applicability on some examples.

**Definition 2.** *The following properties are defined for a search space $S$, variables $x$ and $y$, values $a$ and $b$, and for a set of variables $V$:*

$$\begin{array}{rcl}
\mathit{fixable}(S,x,a) & \equiv & \forall t \in S \;\; (t \in \mathit{Sol}(C) \;\rightarrow\; t[x := a] \in \mathit{Sol}(C)) \\
\mathit{substitutable}(S,x,a,b) & \equiv & \forall t \in S \;\; \left( \begin{array}{l} t_x = a \wedge t \in \mathit{Sol}(C) \;\rightarrow \\ \qquad\qquad t[x := b] \in \mathit{Sol}(C) \end{array} \right) \\
\mathit{removable}(S,x,a) & \equiv & \forall t \in S \;\; \left( \begin{array}{l} t_x = a \wedge t \in \mathit{Sol}(C) \;\rightarrow \\ \qquad \exists b \neq a \;\; (t[x := b] \in \mathit{Sol}(C)) \end{array} \right) \\
\mathit{inconsistent}(S,x,a) & \equiv & \forall t \in S \;\; \left( \; t \in \mathit{Sol}(C) \;\rightarrow\; t_x \neq a \; \right) \\
\mathit{implied}(S,x,a) & \equiv & \forall t \in S \;\; \left( \; t \in \mathit{Sol}(C) \;\rightarrow\; t_x = a \; \right) \\
\mathit{determined}(S,x) & \equiv & \forall t \in S \;\; \left( \begin{array}{l} t \in \mathit{Sol}(C) \;\rightarrow \\ \qquad \forall b \neq t_x \;\; (t[x := b] \notin \mathit{Sol}(C)) \end{array} \right) \\
\mathit{dependent}(S,V,y) & \equiv & \forall t,t' \in S \left( \left( \begin{array}{l} t \in \mathit{Sol}(C) \wedge \\ t' \in \mathit{Sol}(C) \wedge \\ \forall x \in V \;\; (t_x = t'_x) \end{array} \right) \rightarrow t_y = t'_y \right) \\
\mathit{irrelevant}(S,x) & \equiv & \forall t \in S \;\; \left( \begin{array}{l} t \in \mathit{Sol}(C) \;\rightarrow \\ \qquad \forall a \in D \;\; (t[x := a] \in \mathit{Sol}(C)) \end{array} \right)
\end{array}$$

*As for interchangeability, it is well-known (Freuder, 1991) that it can be defined in terms of substitutability:*

$$\mathit{interchangeable}(S,x,a,b) \;\; \equiv \;\; \mathit{substitutable}(S,x,a,b) \;\wedge\; \mathit{substitutable}(S,x,b,a)$$

In the few cases where an ambiguity arises on the considered set of constraints, we will indicate it using subscript (e.g., $\mathit{irrelevant}_C(S,x)$). Note that all the definitions but the last three are *value*-oriented, in that they are properties of specific values of the domain. On the contrary, determinacy, irrelevance, and dependence are *variable-oriented* properties which do not directly express results on particular values of the domains but have important relations with the value-oriented notions (cf. forthcoming Section 2.4).

As already claimed in Section 1, some of the properties of Definition 2 are already known, as well as their beneficial effects to search. In particular, the notion of consistency was proposed by Montanari (1974) and Mackworth (1977), and is one of the best-studied notions in CSP. Substitutability and interchangeability were introduced by Freuder (1991). Implied values, which are also known in the literature as *backbones*, were seemingly first studied explicitly by Monasson, Zecchina, Kirkpatrick, Selman, and Troyansky (1999). To the best of our knowledge, the notions of removable and fixable values (which, as we show in Section 2.4 play a key role in the unifying framework proposed in this paper) have on the contrary not been considered. Determined, irrelevant and dependent variables have been studied in a number of contexts as logic, SAT, and databases, cf., e.g., Beth definability (Chang & Keisler, 1990), *don't care* variables in propositional formulae (Thiffault,

Bacchus, & Walsh, 2004; Safarpour, Veneris, Drechsler, & Lee, 2004), and the *Audit* problem (Jonsson & Krokhin, 2008), but we are aware of little work concerning their application in the context of CSPs.

The following examples illustrate some of the properties.

**Example 3** (Example 2 continued). *Consider a CSP modeling the coloring problem for the graph in Figure 2. Let $\Sigma$ denote the search space in which all five variables $x_1, \ldots, x_5$ have domain $\{R, G, B\}$. The following properties hold:*

- fixable($\Sigma$,$x_1$,$R$), *since for every solution $t$, $t[x_1 := R]$ is a solution as well;*
- substitutable($\Sigma$,$x_1$,$R$,$G$), *since for every solution $t$ such that $t_{x_1} = R$, $t[x_1 := G]$ is a solution as well;*
- interchangeable($\Sigma$,$x_1$,$R$,$G$), *from the previous point and the fact that substitutable($\Sigma$,$x_1$,$G$,$R$) also holds;*
- removable($\Sigma$,$x_1$,$G$), *because for every solution $t$ such that $t_{x_1} = G$, there exists a different color $K \in \{R, B\}$ for $x_1$ such that $t[x_1 := K]$ is a solution as well;*
- irrelevant($\Sigma$,$x_1$), *because we can actually replace the $x_1$-component of any solution $t$ with any other value, since $x_1$ denotes a disconnected node of the graph.*

*The above properties holding for variable $x_1$ that encodes the disconnected node give some initial suggestions on the relationships that exist among the different notions. As for the other nodes, we have, for example:*

- removable($\Sigma$,$x_5$,$G$), *because for every solution $t$ such that $t_{x_5} = G$, there exists a different color $K \in \{R, B\}$ for $x_5$ such that $t[x_5 := K]$ is a solution as well. This is due to the fact that node 5 is connected only to node 4.*

As another, more complex, example, consider the following:

**Example 4.** *Let a CSP be given over boolean variables $x, y, z, w, p, q, r$, whose constraints are encoded by the formula below:*

$$x \;\wedge\; (x \rightarrow y) \;\wedge\; ((z \vee w) \leftrightarrow p) \;\wedge\; ((z \vee y) \leftrightarrow (q \wedge r))$$

*Denoting as $\Xi$ the search space in which all variables range over $\{\mathsf{true}, \mathsf{false}\}$, we have, among the others:*

- inconsistent($\Xi$,$x$,false),
- fixable($\Xi$,$x$,true),
- implied($\Xi$,$x$,true),
- implied($\Xi$,$y$,true),
- inconsistent($\Xi$,$y$,false),
- determined($\Xi$,$y$),
- dependent($\Xi$,$\{z, w\}$,$p$),
- dependent($\Xi$,$\{z, y\}$,$q$),
- dependent($\Xi$,$\{z, y\}$,$r$),
- fixable($\Xi$,$q$,true),

- implied(Ξ,$q$,true),
- fixable(Ξ,$r$,true),
- implied(Ξ,$r$,true).

Some of the definitions of Definition 2 can be used to construct *solution-preserving* transformations, i.e., mappings which transform solutions into solutions.

**Definition 3** (solution-preserving transformation)**.** *A* solution-preserving transformation *is a total mapping $\tau$ from $S$ to $S$ such that*

$$\forall t \in S \ (t \in Sol(C) \rightarrow \tau(t) \in Sol(C))$$

To understand the connection between solution-preserving transformations and the aforementioned properties, consider the following mappings:

$$\tau_1(t) = t[x := a]$$

$$\tau_2(t) = \begin{cases} t[x := b] & \text{if } t_x = a \\ t & \text{otherwise} \end{cases} \qquad \tau_3(t) = \begin{cases} t[x := b] & \text{if } t_x = a \\ t[x := a] & \text{if } t_x = b \\ t & \text{otherwise} \end{cases}$$

Checking whether value $a$ is fixable for variable $x$, whether value $a$ is substitutable to value $b$ for variable $x$, and whether values $a$ and $b$ are interchangeable for variable $x$ amounts to check whether mappings $\tau_1$, $\tau_2$ and $\tau_3$ (respectively) are solution-preserving.

Solution-preserving transformations are interesting because they allow us to remove values from the search space while preserving the *satisfiability* of the problem. Moreover, the correspondence between some properties and the existence of particular solution-preserving mappings shows that interesting connections hold among these properties and other concepts, like symmetries. As an example, Mancini and Cadoli (2005) give a logical characterisation of symmetries in problem specifications, which is very similar to, and in fact stronger than, that of Definition 3. In addition, more general forms of solution-preserving transformations could be defined, that, e.g., allow also for the modification of the constraints, i.e., as pairs $(\tau, \sigma)$ such that $\forall t \in S \ (t \in Sol(C) \rightarrow \tau(t) \in Sol(\sigma(t, C)))$. This interesting topic, that may lead to the definition of further and more general properties of a CSP, is left for future research.

### 2.4 Semantical Relationships

As already observed (cf. also Examples 3 and 4), several semantical relationships exist among the notions presented in Definition 2, some of which appear weaker, while some others stronger. The main connections are clarified by the following theorem.

**Theorem 1.** *The relationships shown in Figure 3 hold between the properties of Definition 2.*

*Proof.*

*Dependence-determinacy.* We have *dependent*$(S, \{x_1, \ldots, x_i\}, y)$ iff every solution $t$ has a value on $y$ which is given by a function $f$ of the values it assigns to $x_1 \ldots x_i$, iff in any search space $\sigma_{x_1=a_1,\ldots,x_i=a_i}(S)$ (where all these variables receive a fixed value), all solutions assign the same value $f(a_1, \ldots, a_n)$ to $y$.

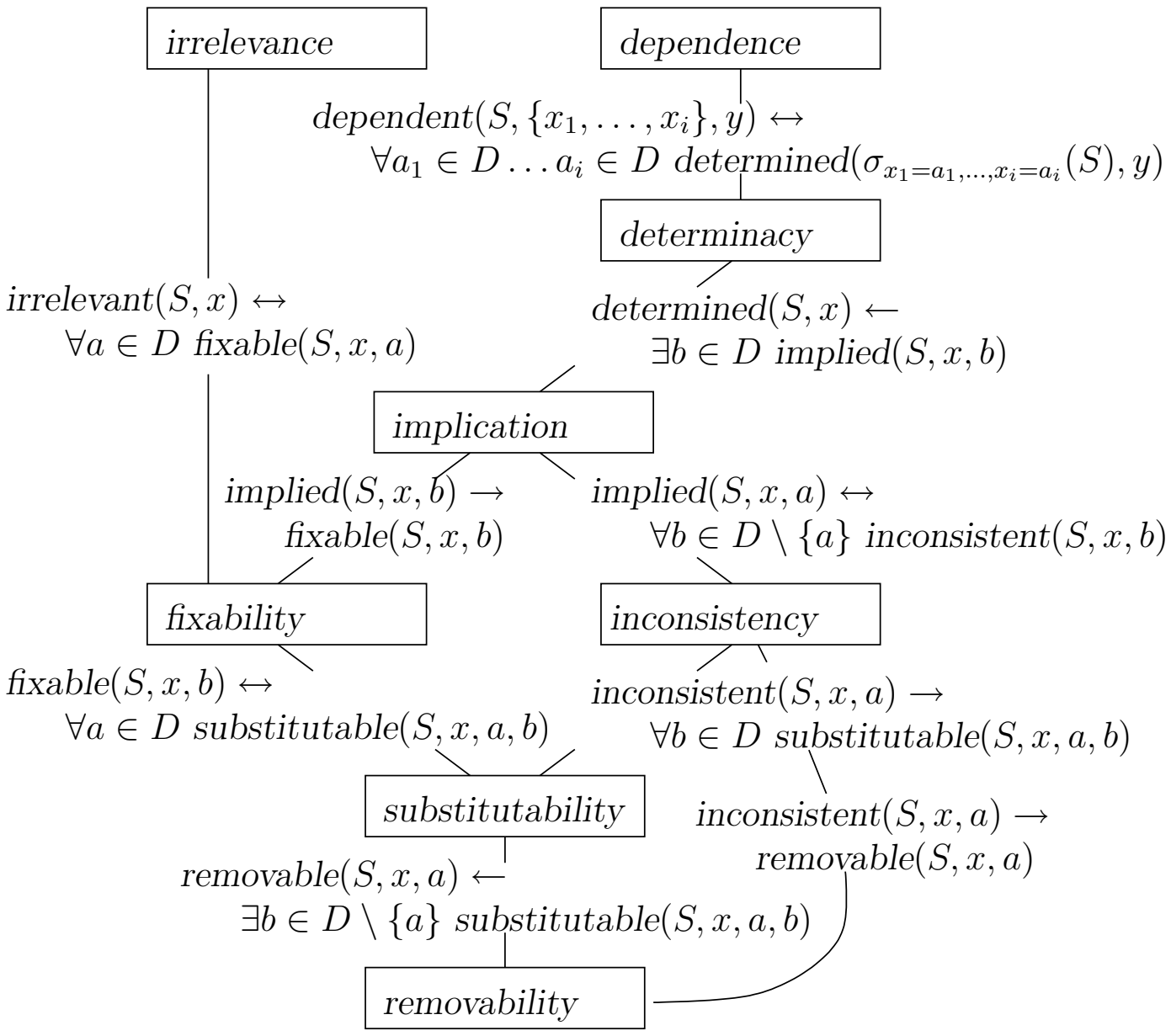

Figure 3: Semantical relationships among the properties.

*Irrelevance-fixability.* $t \in Sol(C) \to \forall a \in D\ (t[x := a] \in Sol(C))$ rewrites to $\forall a \in D\ (t \in Sol(C) \to t[x := a] \in Sol(C))$.

*Determinacy-implication.* If $implied(S, x, b)$ holds for some $b$, then for each $t$ and $a \neq b$ we have $t[x := a] \notin Sol(C)$.

*Implication-fixability.* $implied(S, x, b)$ means that every $t \in Sol(C)$ has $t_x = b$. Hence for every $t \in Sol(C)$, we have $t[x := b] = t \in Sol(C)$.

*Implication-inconsistency.* $implied(S, x, a)$ holds iff $\forall t\ (t_x \neq a \to t \notin Sol(C))$, i.e., iff $\forall t\ \forall b \in D \setminus \{a\}\ (t_x = b \to t \notin Sol(C))$. This rewrites to $\forall b \in D \setminus \{a\}\ inconsistent(S, x, b)$.

*Fixability-substitutability.* Let $D = \{a_1, .., a_d\}$. We have $\bigwedge_{i \in 1..d} substitutable(S, x, a_i, b)$ iff $\forall t\ ((t_x = a_1 \vee \cdots \vee t_x = a_d) \wedge t \in Sol(C) \to t[x := b] \in Sol(C))$, which rewrites to $fixable(S, x, b)$.

*Inconsistency-substitutability.* Suppose $inconsistent(S, x, a)$ holds. No solution $t$ with $t_x = a$ exists, hence the implication $t_x = a \wedge t \in Sol(C) \to t[x := b] \in Sol(C)$ is true for all choices of $b$.

*Inconsistency-removability.* Same argument as for inconsistency-substitutability.

*Substitutability-removability.* Suppose $substitutable(S, x, a, b)$ holds for some value $b \neq a$. This can be written $\exists b\ \forall t\ (t_x = a \wedge t \in Sol(C) \to t[x := b] \in Sol(C))$, which implies $\forall t\ \exists b (t_x = a \wedge t \in Sol(C) \to t[x := b] \in Sol(C))$. The latter rewrites to $\forall t\ (t_x = a \wedge t \in Sol(C) \to \exists b\ t[x := b] \in Sol(C))$. □

Note also that implied values and determined variables are strongly related to problems with a *unique solution*: if a problem has a unique solution, then all its variables have an implied value (cf. Example 1), hence they are determined.

### 2.5 Exploiting Properties in Constraint Solving

An important reason why the aforementioned properties are interesting is that, when detected, they allow us to reduce the search space by removing values from the active domains of the variables. In particular, several properties have successfully been exploited for this purpose, like inconsistency (Montanari, 1974; Mackworth, 1977), substitutability (Freuder, 1991), irrelevance (Thiffault et al., 2004; Safarpour et al., 2004), implication (Monasson et al., 1999), dependence (Mancini & Cadoli, 2007). However, thanks to the unifying framework of Figure 3, we can now show that the wide interest in the aforementioned properties essentially relies in their relations with the two fundamental properties of removability and fixability.

**Theorem 2.** *Let $\pi$ be a CSP $\langle X, D, C\rangle$. If value $a$ is fixable for variable $x \in X$, then $\pi$ is satisfiable if and only if the CSP $\pi' = \langle X, D, C \cup \{x = a\}\rangle$ obtained from $\pi$ by instantiating variable $x$ to value $a$ is satisfiable.*

*Proof.* Assume that $\pi'$ is satisfiable. Then there exists a $X$-tuple $t$ that satisfies all the constraints. Since the constraints of $\pi'$ are a superset of those of $\pi$, $t$ is also a solution of $\pi$.

As for the other direction, assume that $\pi$ is satisfiable. Then there exists a solution $t$. Since value $a$ if fixable for $x$, it follows that $t[x := a]$ is a solution as well. $t[x := a]$ satisfies also the additional constraint "$x = a$" of $\pi'$, hence the latter problem is satisfiable as well. □

**Theorem 3.** *Let $\pi$ be a CSP $\langle X, D, C\rangle$. If value $a$ is removable for variable $x \in X$, then $\pi$ is satisfiable if and only if the CSP $\pi' = \langle X, D, C \cup \{x \neq a\}\rangle$ obtained from $\pi$ by removing value $a$ from the domain of variable $x$ is satisfiable.*

*Proof.* If $\pi'$ is satisfiable, then, by the same arguments of the proof of Theorem 2, $\pi$ is satisfiable as well.

As for the other direction, assume that $\pi$ is satisfiable. Then there exists a solution $t$. If $t_x \neq a$, $t$ is of course also a solution of $\pi'$. On the other hand, if $t_x = a$, since value $a$ is removable for $x$ in $\pi$, it follows that there exists $b \neq a$ such that $t[x := b]$ is a solution as well. $t[x := b]$ satisfies also the additional constraint "$x \neq a$" of $\pi'$, hence the latter problem is also satisfiable. □

The above results show the key roles played by fixability and removability. They are the *ideal* properties that should be checked in order to reduce the domain for a variable. The interest in the other notions essentially relies on their relationships with fixability and removability. As an example, an implied value is of interest essentially because it is fixable, an irrelevant variable is of interest essentially because it is fixable to any value of its domain, a substitutable value is of interest essentially because it is removable, etc.

Also the interest in inconsistent values relies on the fact that they are removable. However, inconsistency is much stronger that removability, because removing inconsistent values guarantees that *all* solutions (and not only the satisfiability of the problem) are preserved. Hence removability plays exactly the same role of inconsistency in case we do not want to find all solutions of a problem but simply want to find one. In such situations removability is the ideal property to use.

As for the other properties, it is worth noting that, although they may appear very strong and unlikely at a first sight, they can still play a precious role when detected dynamically during search. As an example, Thiffault et al. (2004) show that dynamically detecting variables that become irrelevant during search (called *don't care* variables in the paper) can greatly speed-up non-CNF SAT solvers, by actually separating problems into independent components.

## 3. Complexity of reasoning

In this section we show that the problem of checking whether properties of Definition 2 hold is coNP-complete. Hence, in Section 3.2, we try to determine special cases where checking can be done efficiently (i.e., in polynomial time).

### 3.1 Intractability Results

From now on, we assume that the input is given as a set of constraints $C$ over a set of variables $X$. We also assume that the problem of checking whether $t \in Sol(C)$ is polynomial in the size of the representation of the input. Such properties hold for propositional logic and for CSPs, in the sense of Dechter (1992).

We note that the problem of checking whether properties of Definition 2 hold is in coNP, because, for each of them it is possible to find a counter-example by guessing a tuple in $S$ (two, in case of dependency) in non-deterministic polynomial time, and checking, in polynomial time, whether the negation of the subformula between parentheses holds (as for interchangeability, we note that the logical "and" of two properties in coNP is still in coNP). Alternatively, coNP-membership follows observing that succinct certificates exist proving that the various properties do not hold (as an example, a certificate proving that variable $x$ is not fixable to $a$ is a $V$-tuple $t \in Sol(C)$ such that $t[x := a] \notin Sol(C)$). In the rest of this section, proofs are therefore restricted to the coNP-hardness part.

**Theorem 4** (coNP-completeness of properties of Definition 2)**.** *Given a CSP, the following tasks are coNP-complete:*

- *Checking whether value $a$ is fixable, removable, inconsistent, implied, or determined for variable $x \in X$;*
- *Checking whether value $a \in D$ is substitutable to, or interchangeable with value $b \in D$ for variable $x \in X$;*
- *Checking whether variable $y \in X$ is dependent on variables in a set $V \subseteq X$;*
- *Checking whether variable $x \in X$ is irrelevant.*

*Proof.* To prove that checking such properties is hard for coNP, we reduce a coNP-complete problem, i.e., that of checking that an arbitrary CSP is unsatisfiable, to the problem of checking these properties. In particular, the proofs hold even if the domain is boolean, in which case the CSP can be written as a propositional formula $\phi$, e.g., in CNF. Hence, let $\phi$ be an arbitrary propositional formula in CNF over set of variables $X$, and let $x$ be a variable such that $x \notin X$: the unsatisfiability problem of $\phi$ can be reduced into the problem of checking the various properties. Moreover, the semantical relationships defined in Theorem 1 allow to infer in an elegant way hardness results for several properties starting from those of the others.

*Irrelevance.* Consider $\psi$ defined as $\phi \wedge x$. $\psi$ is unsatisfiable if and only if $\phi$ is unsatisfiable. We now show that $\phi$ is unsatisfiable if and only if $x$ is irrelevant in formula $\psi$. Let us first assume that $\phi$ is unsatisfiable. It follows that $x$ is irrelevant in $\psi$, because $\psi$ has no models. On the other hand, let $M$ be a model of $\phi$. Interpretation $M \cup \{x \leftarrow \mathit{true}\}$ is a model of $\psi$, while $M \cup \{x \leftarrow \mathit{false}\}$ is not, implying that $x$ is not irrelevant in $\psi$.

*Fixability, Substitutability.* Results follow from that of irrelevance, combined with the semantical relationships that define irrelevance in terms of fixability, and fixability in terms of substitutability.

*Dependence.* Consider $\psi$ defined as $\phi \wedge (x \vee \neg x)$. $\psi$ is unsatisfiable if and only if $\phi$ is unsatisfiable, and that $x$ is dependent on $X$ in $\psi$ if and only if $\phi$ is unsatisfiable.

*Determinacy.* The result follows from that of dependence, combined with the semantical relationship that defines dependence in terms of determinacy.

*Implication.* Consider $\psi$ defined as $\phi \wedge x$. Value *false* is implied for $x$ in $\psi$ if and only if $\phi$ is unsatisfiable.

*Inconsistency.* The result follows from that of implication, combined with the semantical relationship that defines implication in terms of inconsistency.

*Removability.* Consider $\psi$ defined as $\phi \wedge x$. Value *true* is removable for $x$ in $\psi$ if and only if $\phi$ is unsatisfiable.

□

From the proof of Theorem 4, it can be observed that the intractability of checking each of the properties holds also for binary CSPs (i.e., CSPs in which all constraints relate at most two variables).

**Theorem 5** (coNP-completeness of properties of Definition 2 for binary CSPs)**.** *Given a CSP with only binary constraints on a domain of size greater than two, checking the properties of Definition 2 is coNP-complete.*

*Proof.* We give the proof for irrelevance only: the others can be derived similarly.

Let $\Phi = \langle X, D, C \rangle$ be a binary CSP. Consider an arbitrary variable $x \notin X$ and let $a$ and $b$ be arbitrary distinct values in $D$. Let $\Psi$ denote the CSP $\langle X', D, C' \rangle$ with $X' = X \cup \{x\}$, and $C' = C \cup \{x = a\}$. $\Psi$ is binary and, similarly to the proof of Theorem 4, $\Phi$ is unsatisfiable if and only if variable $x$ is irrelevant for $\Psi$.

From the observation that a CSP encoding of the graph 3-colourability problem can be made using only binary constraints, the thesis follows, since checking unsatisfiability of this problem (which is coNP-hard) can be reduced to checking irrelevance in a binary CSP. □

### 3.2 Tractability Results

Since detecting any of the properties we are interested in is a computationally hard problem, a natural question is to determine special cases where checking can be done efficiently. To this end, we investigate two approaches: we exhibit *syntactical restrictions* which make the problem tractable, and we study *local* relaxations of the definitions which are polynomial-time checkable, and which therefore provide incomplete algorithms for detecting the various properties.

#### 3.2.1 Tractability for Restricted Constraint Languages

A number of syntactical restrictions to the constraint satisfaction problem are known which make it tractable. For instance, in the case of boolean constraints, i.e., propositional formulae, the satisfiability problem becomes tractable if the instance is expressed using only Horn clauses, only dual Horn clauses (i.e., clauses with at most one negative literal), only clauses of size at most 2, or only affine constraints (i.e., formulae built using XOR). These are known as the *Schaefer's (1978) classes.* It is natural to wonder if all the properties identified in Definition 2 are also easy to determine for these classes of formulae. This is indeed the case for most of them, and we give a more general condition under which tractable classes for the consistency property are also tractable for other properties of our framework. We note that a recent paper (Jonsson & Krokhin, 2004) gives a complete characterisation of tractable cases for a related property.

In what follows we are interested in *classes* of CSPs. To this end, we define a *constraint language* $\Gamma_D$ over domain $D$ to be a finite set of relations (of finite arity) with elements in $D$, and denote by CSP($\Gamma_D$) the set of CSPs of the form $\langle X, D, C\rangle$ with every element $c_i = r_i(V_i) \in C$ being such that $V_i \subseteq X$ and $r_i \in \Gamma_D$. (We observe that once the constraint language $\Gamma_D$ is fixed, the domain $D$ for all instances of CSP($\Gamma_D$) is fixed as well.)

A constraint language $\Gamma$ is said to be *closed under instantiation* (resp. *under complementation*) if whenever a constraint $c_i = r_i(V_i)$ is expressible in the language (i.e., $r_i \in \Gamma$), the relation $\pi_{X\setminus\{x\}}(\sigma_{x=a}(c_i)), a \in D$ (resp. the complementation $\overline{c_i}$) can be represented by a *conjunction* of constraints of the language. This means that there exist constraints $c'_1 = r'_1(V'_1), \ldots, c'_k = r'_k(V'_k)$, with $V'_j \subseteq X$ and $r'_j \in \Gamma$ for each $j$, such that $\pi_{X\setminus\{x\}}(\sigma_{x=a}(c_i))$ (resp. $\overline{c_i}$) is equivalent to $c'_1 \wedge \cdots \wedge c'_k$.[2] Well known constraint languages on boolean domains which are closed under instantiation and complementation are those of Horn clauses, dual Horn clauses, 2CNF clauses or affine constraints (since the instantiation or the complement of a Horn /dual Horn/2CNF clause or affine constraint can be expressed as a conjunction of Horn /dual Horn/2CNF clauses or affine constraints).

**Theorem 6.** *Given a constraint language $\Gamma$, if the satisfiability problem CSP($\Gamma$) is tractable and if $\Gamma$ is closed under instantiation, then the problem of checking determinacy for CSPs in the class CSP($\Gamma$) is tractable as well.*

2. Note that we define closure with respect to *complementation* with a slightly different non-standard meaning, as the negation of the constraint needs be expressible as a *conjunction* of constraints. Some definitions impose that it be definable as a *single* constraint of the language.

*Proof.* Let us consider an arbitrary instance $\langle X, D, C\rangle \in \mathrm{CSP}(\Gamma)$. Variable $x \in X$ is not determined if and only if there exist two different domain values, $a$ and $b \in D$, such that

$$\pi_{X\setminus\{x\}}(\sigma_{x=a}(Sol(C))) \cap \pi_{X\setminus\{x\}}(\sigma_{x=b}(Sol(C)))$$

is not empty, i.e., if and only if one of the CSPs $\langle \hat{X}, D, \hat{C}_{a,b}\rangle$, with:

- $\hat{X} = X \setminus \{x\}$, and
- $\hat{C}_{a,b} = \{\pi_{X\setminus\{x\}}(\sigma_{x=a}(c)) \mid c \in C\} \cup \{\pi_{X\setminus\{x\}}(\sigma_{x=b}(c)) \mid c \in C\}$,

is satisfiable. If $\Gamma$ is closed under instantiation, constraints in $\hat{C}_{a,b}$ can all be written as conjunctions of constraints in $\Gamma$. Hence, we have reduced the determinacy-testing problem to solving $O(|D|^2)$ instances of $\mathrm{CSP}(\Gamma)$, which is tractable. □

**Theorem 7.** *Given a constraint language* $\Gamma$*, if the satisfiability problem CSP(*$\Gamma$*) is tractable and if* $\Gamma$ *is closed under instantiation and complementation, then the problem of checking any property among fixability, substitutability, interchangeability, inconsistency or irrelevance for CSPs in the class CSP(*$\Gamma$*) is tractable as well.*

*Proof.* We start with the substitutability property and note that, given an arbitrary instance $\langle X, D, C\rangle \in \mathrm{CSP}(\Gamma)$, value $a \in D$ is substitutable by $b \in D$ for variable $x \in X$ if and only if

$$\pi_{X\setminus\{x\}}(\sigma_{x=a}(Sol(C))) \quad \subseteq \quad \pi_{X\setminus\{x\}}(\sigma_{x=b}(Sol(C))).$$

This inclusion is false, i.e., substitutability does *not* hold, if and only if the set

$$\pi_{X\setminus\{x\}}(\sigma_{x=a}(Sol(C))) \quad \cap \quad \overline{\pi_{X\setminus\{x\}}(\sigma_{x=b}(Sol(C)))} \tag{1}$$

is non-empty. Since $\sigma_{x=b}(Sol(C)) \;=\; \sigma_{x=b}(\bigcap_{c\in C} Sol(c)) \;=\; \bigcap_{c\in C}(\sigma_{x=b}(Sol(c)))$, we have:

$$\pi_{X\setminus\{x\}}(\sigma_{x=b}(Sol(C))) \;=\; \pi_{X\setminus\{x\}}\left(\ \textstyle\bigcap_{c\in C}(\sigma_{x=b}(Sol(c)))\ \right).$$

Although the projection of an intersection of relations is not equal to the intersection of their projections in general, the latter rewrites to:

$$\bigcap_{c\in C} \pi_{X\setminus\{x\}}(\sigma_{x=b}(Sol(c))).$$

This is due to the fact that we select on $x$ before eliminating it by projection. We only prove the inclusion which does not hold in general: suppose we have $t \in \bigcap_{c\in C} \pi_{X\setminus\{x\}}(\sigma_{x=b}(Sol(c)))$. This means that $\forall c \in C$, there exists a tuple $t^c$ such that $t^c|_{X\setminus\{x\}} = t$ and $t^c \in \sigma_{x=b}(Sol(c))$. It follows that $t^c_x = b$ and that we have indeed a unique $t$ with $t_x = b$ and $t^c|_{X\setminus\{x\}} = t$ which is such that $\forall c \in C$ $(t \in \sigma_{x=b}(Sol(c)))$, i.e., $t \in \pi_{X\setminus\{x\}}(\bigcap_{c\in C}(\sigma_{x=b}(Sol(c))))$.

Formula (1) is therefore equivalent to:

$$\bigcap_{c\in C} \pi_{X\setminus\{x\}}(\sigma_{x=a}(Sol(c))) \quad \cap \quad \bigcup_{c\in C} \overline{\pi_{X\setminus\{x\}}(\sigma_{x=b}(Sol(c)))}$$

A solution exists (and therefore substitutability does not hold) if one of the sets

$$\bigcap_{c \in C} \pi_{X \setminus \{x\}}(\sigma_{x=a}(Sol(c))) \quad \cap \quad \overline{\pi_{X \setminus \{x\}}(\sigma_{x=b}(Sol(c)))}$$

obtained for every $c \in C$ has a solution. If language $\Gamma$ is closed under instantiation and complement, we can express the new constraint $\overline{\pi_{X \setminus \{x\}}(\sigma_{x=b}(Sol(c)))}$ as a conjunction $C'$ of constraints in $\Gamma$. Each of the sets has a solution iff the CSP $\langle X \setminus \{x\}, D, \{\pi_{X \setminus \{x\}}(\sigma_{x=a}(c)) \mid c \in C\} \cup C' \rangle$ is satisfiable. We have reduced the substitutability testing problem to solving $|C|$ instances of a constraint satisfaction problem whose constraints are all in the language $\Gamma$, which is tractable.

The results for fixability, interchangeability and irrelevance follow directly, because of the semantical relationships shown in Figure 3. Consistency of value $a$ for variable $x$ can directly be expressed as the satisfiability of $\pi_{X \setminus \{x\}}(\sigma_{x=a}(Sol(C)))$, which can be expressed in $\Gamma$, and the proof for implication follows from this result. □

A slightly different closure property is needed for the removability of value $a$ for variable $x$ since it is expressed as $\pi_{X \setminus \{x\}}(Sol(C)) \quad \subseteq \quad \pi_{X \setminus \{x\}}(\sigma_{x \neq a}(Sol(C)))$.

Nevertheless, since on boolean domains a value $v$ is *removable* if $v$ is substitutable by $\neg v$, and from the remarks on the closure properties of Schaefer's classes and the previous theorem, we obtain that:

**Corollary 1.** *Testing fixability, substitutability, interchangeability, inconsistency, determinacy, irrelevance and removability is tractable for a boolean CSP where constraints are either Horn clauses, dual Horn clauses, clauses of size at most two or affine constraints.*

Unfortunately, we don't have tractability results for dependence.

Table below summarizes Theorems 6 and 7, and Corollary 1:

| Property | Polynomial if $\Gamma$ is |
|---|---|
| Determinacy | Tractable and closed under instantiation |
| Fixability, substitutability, interchangeability, inconsistency, irrelevance | Tractable and closed under inst. and compl. |
| Removability | Boolean Schaefer |

It is worth noting that conditions become more restrictive when reading the table top-down.

Moreover, in all the cases, it can be observed that tractability of the various property-detection problems derives from tractability of the constraint language $\Gamma$. This leads to serious concerns about the practical applicability of such results: if CSP($\Gamma$) is tractable, why worrying for identifying such properties? Actually, preliminary studies show that better results are unlikely to hold: as an example, it can be proven that if the constraint language is intractable, then there is *no hope* for detecting properties like fixability, irrelevance, substitutability, inconsistency in polynomial time. So these results become of interest, suggesting two main directions of further research: the first is of course that of investigating the *practical* benefit of detecting such properties in real cases; the second is to exploit sufficient but efficiently evaluable conditions for these properties to hold, that can be regarded as a form of *approximate* reasoning. One of the most used forms of such kind of reasoning is called *local reasoning*, which is addressed below.

#### 3.2.2 Tractability Through Locality

An important class of incomplete criteria to determine in polynomial time whether a complex property holds are those based on *local* reasoning. This approach has proved extremely successful for consistency (Mackworth, 1977) and interchangeability (Freuder, 1991) (cf. also Choueiry & Noubir, 1998, where a classification of different local forms of interchangeability are studied and classified). We propose in this section a systematic investigation of whether a local approach can be used for *value-based* properties.

Verifying a property $P(C)$ of a set of constraints $C$ *locally* means that we verify the property on a well-chosen number of sub-problems. We must ensure that this approach is sound for the considered property:

**Definition 4** (soundness of local reasoning)**.** *We say that local reasoning on a property $P$ is* sound *if, for all subsets of constraints $C_1 \subseteq C, \ldots, C_k \subseteq C$ such that $\bigcup_{i \in 1..k} C_i = C$, we have (depending on the property):*

$$\left( \bigwedge_{i \in 1..k} P(C_i) \right) \rightarrow P(C) \quad \text{or} \quad \left( \bigvee_{i \in 1..k} P(C_i) \right) \rightarrow P(C).$$

A typical choice of granularity is to simply consider that each $C_i$ contains one of the constraints of $C$ as is done, for instance, for arc-consistency. On the other extreme, if we take a unique $C_1 = C$, we have a global checking. Between these two extremes, a wide range of intermediate levels can be defined (cf., e.g., Freuder, 1978, 1991).

**Example 5.** *Consider the CSP $\langle X, D, C\rangle$ with $X = \{x, y, z\}$, $D = \{0, 1, 2\}$ and $C = \{c_1, c_2, c_3\}$, whose elements are defined as follows:*

| $x$ | $y$ |
|---|---|
| 0 | 1 |
| 1 | 2 |
| 2 | 1 |
| 2 | 2 |

$c_1(x, y)$

| $x$ | $z$ |
|---|---|
| 1 | 0 |
| 1 | 2 |
| 2 | 0 |
| 2 | 2 |

$c_2(x, z)$

| $y$ | $z$ |
|---|---|
| 1 | 1 |
| 1 | 2 |
| 2 | 1 |
| 2 | 2 |

$c_3(y, z)$

*It can be observed that value 1 is substitutable to 2 for variable $x$. In order to check this property locally, we consider a suitable covering $C_1, \ldots, C_k$ of the set $C$, and verify it on the induced subproblems. As an example, by taking $C_1 = \{c_1\}$, $C_2 = \{c_2\}$, $C_3 = \{c_3\}$, we have $substitutable_{C_i}(S, x, 1, 2)$ for every $i \in \{1, 2, 3\}$. Since local reasoning is* sound *for substitutability (cf. Freuder, 1991), we can infer that the global property $substitutable_C(S, x, 1, 2)$ holds.*

Reasoning locally is typically tractable if we focus on a moderate number of subsets of $C$, and under the condition that we can bound the complexity of reasoning on each of these subsets. A typical assumption in CSP is that we can bound the arity of the constraints, and that every constraint is for instance binary. In this case, the cost of determining any property of the constraint is polynomial; and if we choose to reason locally by considering each constraint separately, or by taking groups of constraints of bounded size, then local checking is tractable.

**Theorem 8.** *Local reasoning is sound for the properties of substitutability, interchangeability, fixability, inconsistency, implication, irrelevance, determinacy, dependence. In particular, for any* $C_1 \subseteq C, \ldots, C_k \subseteq C$ *such that* $\bigcup_{i \in 1..k} C_i = C$*:*

- $\left( \bigwedge_{i \in 1..k} \mathit{substitutable}_{C_i}(S, x, a, b) \right) \to \mathit{substitutable}_{C}(S, x, a, b)$*;*
- $\left( \bigwedge_{i \in 1..k} \mathit{interchangeable}_{C_i}(S, x, a, b) \right) \to \mathit{interchangeable}_{C}(S, x, a, b)$*;*
- $\left( \bigwedge_{i \in 1..k} \mathit{fixable}_{C_i}(S, x, b) \right) \to \mathit{fixable}_{C}(S, x, b)$*;*
- $\left( \bigvee_{i \in 1..k} \mathit{inconsistent}_{C_i}(S, x, a) \right) \to \mathit{inconsistent}_{C}(S, x, a)$*;*
- $\left( \bigvee_{i \in 1..k} \mathit{implied}_{C_i}(S, x, a) \right) \to \mathit{implied}_{C}(S, x, a)$*;*
- $\left( \bigwedge_{i \in 1..k} \mathit{irrelevant}_{C_i}(S, x) \right) \to \mathit{irrelevant}_{C}(S, x)$*;*
- $\left( \bigvee_{i \in 1..k} \mathit{determined}_{C_i}(S, x) \right) \to \mathit{determined}_{C}(S, x)$*;*
- $\left( \bigvee_{i \in 1..k} \mathit{dependent}_{C_i}(S, V, y) \right) \to \mathit{dependent}_{C}(S, V, y)$*.*

*Proof.* The result is well-known for consistency (Mackworth, 1977), substitutability and interchangeability (Freuder, 1991). Fixability of variable $x$ to value $b$ can be expressed as

$$\forall a \neq b \; (\mathit{substitutable}_{C}(S, x, a, b))$$

Therefore, if we have $\bigwedge_{i \in 1..k} \mathit{fixable}_{C_i}(S, x, b)$ (which is equivalent to $\bigwedge_i \bigwedge_{a \neq b} \mathit{substitutable}_{C_i}(S, x, a, b)$ and to $\bigwedge_{a \neq b} \bigwedge_i \mathit{substitutable}_{C_i}(S, x, a, b)$), then we have $\bigwedge_{a \neq b} \mathit{substitutable}_{C}(S, x, a, b)$, which means $\mathit{fixable}_{C}(S, x, b)$. A similar argument works for irrelevance, which can be analogously defined in terms of fixability (cf. Figure 3). As for implication, if a value $a$ is implied for variable $x$ in any $C_i$, then all tuples $t$ with $t_x \neq a$ violate the constraints of $C_i$ and do *a fortiori* not belong to $\mathit{Sol}(C)$. Similarly for determinacy: if a variable $x$ is determined in any $C_i$, then all tuples $t \in \mathit{Sol}(C_i)$ will be such that $t[x := b] \notin \mathit{Sol}(C_i)$ for any $b \neq t_x$. Hence there cannot be any solution $t \in \mathit{Sol}(C)$ such that $t[x := b] \in \mathit{Sol}(C)$ with $b \neq t_x$. Finally, as for dependence, if there exists a $C_i$ for which $\mathit{dependent}_{C_i}(S, V, y)$ holds (it is enough to consider sets of constraints $C_i$s such that $V \cap \mathit{var}(C_i) \neq \emptyset$ and $y \in \mathit{var}(C_i)$ for which we have $\mathit{dependent}_{C_i}(S, V \cap \mathit{var}(C_i), y)$), we have, by definition, that $\forall t, t' \in \mathit{Sol}(C_i) \;\; (\forall x \in V \cap \mathit{var}(C_i) \; (t_x = t'_x)) \to t_y = t'_y$, and so $\mathit{dependent}_{C}(S, V, y)$ also holds, since any solution of the whole problem must satisfy also $C_i$. □

**Example 6** (Example 5 continued)**.** *Value 2 is fixable for z. This can be inferred by performing local reasoning as follows:*

- *fixable*$_{C_1}(S, z, 2)$ *holds, since* $z$ *does not occur in the scope of* $c_1$*;*
- *fixable*$_{C_2}(S, z, 2)$ *holds, since for any tuple* $t$ *that belongs to* $c_2$*, also* $t[z := 2]$ *belongs to* $c_2$*;*
- *fixable*$_{C_3}(S, z, 2)$ *holds, because of the same argument.*

*Since by Theorem 8 local reasoning is sound for fixability, we can infer that fixable*$_C(S, z, 2)$ *holds.*

There is only one (value-based) property, namely removability, for which the local approach is unfortunately not sound:

**Theorem 9.** *Local reasoning is **not** sound for the removability property.*

*Proof.* Take $C = C_1 \wedge C_2$, where $C_1$ is defined as $x \leq y$ and $C_2$ as $x \geq y$. Suppose the domain has values $\{1, 2, 3\}$. Value 2 for $x$ is removable from both constraints considered independently since, in both cases, we can change the value of any solution which assigns 2 to $x$ to another value. Still, value 2 is not removable for $x$ from their conjunction. To see why, consider the solution with $\langle x, y\rangle = \langle 2, 2\rangle$, and observe that neither $\langle 1, 2\rangle$ or $\langle 3, 2\rangle$ are solutions.

Note that removing values which are shown to be removable only locally can even make a satisfiable problem unsatisfiable: if furthermore we add the constraints $C_3$, defined as $x \neq 1$ and $C_4$, defined as $x \neq 3$, then value 2 for $x$ is removable in each constraint, while the only (global) solution actually assigns value 2 to $x$. □

This result, although negative, is in fact interesting, because it gives an *ex-post* justification of the extensive use that has been made in the last decades of stronger notions, like inconsistency or substitutability, both of which imply removability (cf. Figure 3). The main reason why current tools and frameworks for CP try to detect these properties is in order to *remove* values from the active domain of some variables. This naturally relies on the removability property (cf. Section 2.5). However, the reason why removability is not directly used is because it is intractable. For that reason, stronger notions like consistency or substitutability are the forms of removability that have been more commonly used. Actually, unlike full-fledged removability (cf. Theorem 9), these properties can be detected efficiently, but incompletely, through local reasoning. Hence, this raises an interesting open issue: do there exist new (i.e., other than substitutability and inconsistency) properties for which local reasoning is sound and which imply removability?

We end this section by noting that the local version of the fixability property is indeed a generalisation to arbitrary domains of the *pure literal rule* (Davis & Putnam, 1960) which is well-known in the case of boolean constraints in conjunctive normal form. The pure literal rule exploits the cases where no constraint (clause) of the problem has a positive (resp. negative) occurrence of some variable $x$. In this case, assigning value 0 (resp. 1) to $x$ preserves the satisfiability of the problem: if a solution $t$ with $t_x = 1$ exists, then $t[x := 0]$ will also be a solution since no clause constrains $x$ to have value 1.

**Example 7.** *Consider the following propositional formula $\varphi$ in CNF:*

$$(x \vee y \vee z) \wedge (x \vee \neg y \vee \neg z) \wedge (y \vee \neg z)$$

*Since $\neg x$ does not occur in any clause, we can assign $x$ to 1, and maintain the satisfiability of the formula: for any solution $t$ of $\varphi$, the assignment $t[x := 1]$ is a solution of $\varphi$ as well.*

It is clear that the pure literal rule detects fixability based on a reasoning local to each clause (a variable $x$ is fixable to, say, 1 in a clause iff this clause does not contain the literal $\neg x$, and the pure literal rule checks that this condition holds for every constraint).

No generalisation of the pure literal rule has, to the best of our knowledge, been proposed for CSP, while a generalisation of the pure literal rule for QBF has been applied to solvers for Quantified CSP under the name *pure value rule* (Gent, Nightingale, & Stergiou, 2005). It can be observed that such proposal is in fact a local relaxation of the generalisation to quantified constraints of fixability. As will be shortly discussed in Section 4, all the properties presented in this paper can be generalized to Quantified CSP in an elegant way, and many local relaxations remain valid.

## 4. Conclusions and Perspectives

In this paper we reconsidered most of the structural properties of CSPs extensively studied and exploited in order to simplify search. Such properties may be of course useful also in other tasks, e.g., for the classification or update of solutions, or for compacting the solution space, and for supporting explanation and interaction with users.

We provided a unifying framework for the properties, that clarifies their semantical relationships and allows new ones to emerge. We argued that some of the new notions, namely *fixability* and *removability* play a key role in a deep understanding of the topic, being the ideal characterisations of values that can be fixed or removed preserving the satisfiability of the problem. Known properties, like inconsistency and substitutability are thus suitable specialisations of them.

We then tackled the questions related to the automated detection of the different properties and of their exploitation by the solving engine for simplifying problems. In particular, we showed how detecting each of the proposed properties is in general as hard as the original CSP. Hence, in order to find efficient ways for their verification, we investigated, according to the CSP approach, two main lines: addition of suitable restrictions of the constraint language and approximation of the reasoning task by exploiting local versions of the various notions. Moreover, we discussed how in some cases such properties may arise from explicit promises made by users. This is the case of problems with properties such as functional dependencies and unique solutions.

Two of the perspectives raised by our work concern the new central properties which have emerged from it. We have identified the removability property as an ideal characterisation of the values which can be removed while preserving satisfiability. Unfortunately, negative results (coNP-completeness of the detection of this property and impossibility of local reasoning) make it impossible to directly use the removability property in practice. This gives an *ex-post* justification for the extensive use that has been made in the last decades of stronger notions (like inconsistency or substitutability) which imply removability, yet can be checked by tractable means (of course at the price of losing completeness). An interesting problem is thus to determine new cases where removability-checking is tractable.

Also, the benefits of fixability have long been known in the boolean case, since this property has been used in the form of the pure literal rule in many SAT solvers. However, no generalisation of this property to CSPs has been considered so far.

Finally, the proposed framework allows a natural and elegant generalisation to the case of Quantified CSP. In particular, in related work (Bordeaux, Cadoli, & Mancini, 2008) we propose the new notion of *outcome* as the natural counterpart at the quantified level of the concept of *solution* of a CSP. With such notion in mind, all the properties studied in this paper can be straightforwardly restated for Quantified CSP, as well as their local relaxations, and new, even more general concepts emerge (the so-called *shallow* properties, that may have an impact also at the pure existential –CSP– level). This opens important new horizons, allowing QCSP solvers to perform a smarter reasoning on the input problem, by taking into proper account the quantifiers prefix, that today is usually ignored.

## Acknowledgments

This paper is an extended and revised version of Bordeaux, Cadoli, and Mancini (2004).